\documentclass[letterpaper]{article} % DO NOT CHANGE THIS
\usepackage{aaai23}  % DO NOT CHANGE THIS
\usepackage{times}  % DO NOT CHANGE THIS
\usepackage{helvet}  % DO NOT CHANGE THIS
\usepackage{courier}  % DO NOT CHANGE THIS
\usepackage[hyphens]{url}  % DO NOT CHANGE THIS
\usepackage{graphicx} % DO NOT CHANGE THIS
\usepackage{natbib}  % DO NOT CHANGE THIS AND DO NOT ADD ANY OPTIONS TO IT
\usepackage{caption} % DO NOT CHANGE THIS AND DO NOT ADD ANY OPTIONS TO IT
\usepackage{algorithm}

\usepackage{algpseudocode}
\usepackage{subcaption}
\usepackage{bm}
\usepackage{amsfonts}       % blackboard math symbols
\usepackage{amsmath}       % blackboard math symbols
\usepackage{booktabs}       % professional-quality tables
\usepackage{array}

\usepackage{newfloat}
\usepackage{listings}
\DeclareCaptionStyle{ruled}{labelfont=normalfont,labelsep=colon,strut=off} % DO NOT CHANGE THIS
\floatstyle{ruled}
\newfloat{listing}{tb}{lst}{}
\floatname{listing}{Listing}
\title{Leveraging Old Knowledge to Continually Learn New Classes in Medical Images}
\author {
    Evelyn Chee\textsuperscript{\rm 1,2},
    Mong Li Lee\textsuperscript{\rm 1,2},
    Wynne Hsu\textsuperscript{\rm 1,2}
}
\affiliations {
    \textsuperscript{\rm 1} School of Computing, National University of Singapore \\
    \textsuperscript{\rm 2} Institute of Data Science, National University of Singapore \\
    \{echee, leeml, whsu\}@comp.nus.edu.sg
}
\begin{document}

\maketitle

\begin{abstract}
Class-incremental continual learning is a core step towards developing artificial intelligence systems that can continuously adapt to changes in the environment by learning new concepts without forgetting those previously learned. This is especially needed in the medical domain where continually learning from new incoming data is required to classify an expanded set of diseases. In this work, we focus on how old knowledge can be leveraged to learn new classes without catastrophic forgetting. We propose a framework that comprises of two main components: (1) a dynamic architecture with expanding representations to preserve previously learned features and accommodate new features; and (2) a training procedure alternating between two objectives to balance the learning of new features while maintaining the model’s performance on old classes. Experiment results on multiple medical datasets show that our solution is able to achieve superior performance over state-of-the-art baselines in terms of class accuracy and forgetting.
\end{abstract}

\section{Introduction}

\begin{figure}[!tb]
	\centering
	\includegraphics[width=0.95\linewidth]{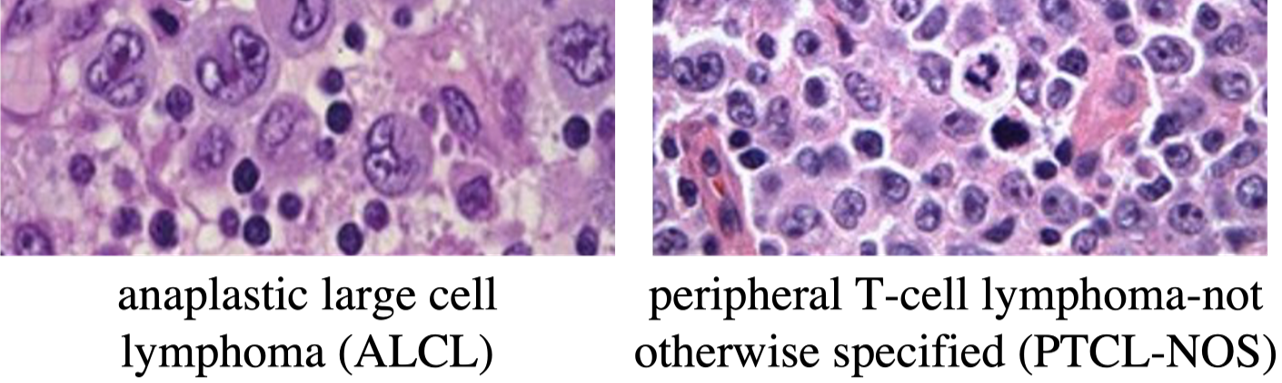}
	\caption{Sample pathology images of different lymphomas \cite{el2012pediatric}.}
	\label{fig:motivation}
\end{figure}

Deep neural networks (DNNs) have excelled in many machine learning classification tasks and shown to achieve human-level performance in medical imaging applications \cite{mckinney2020international,ardila2019end,esteva2017dermatologist}. However, this is under the condition that all classes are known prior to training. This assumption is often violated in the medical domain as some diseases are rare and discovering new disease is not uncommon. In both cases, not all classes would have readily available data for training the DNN. 
Figure \ref{fig:motivation} shows samples of two types of lymphoma with the latter being a rarer disease \cite{el2012pediatric}. A system that has only learn to diagnose the former can easily misclassify the other as the same type with high confidence due to their similar appearance. This can have dire consequences as the treatment strategies for the two are different and initiating the correct regimen is crucial to achieve good outcome. Hence, any DNN systems in the medical domain must be able to continually learn an expanding set of classes as and when new data becomes available. Further, they should do so without negatively affecting the performance for diagnosing previously seen diseases. 

\begin{figure}[!tb]
	\centering
	\begin{subfigure}{\linewidth}
		\centering
		\includegraphics[width=\linewidth]{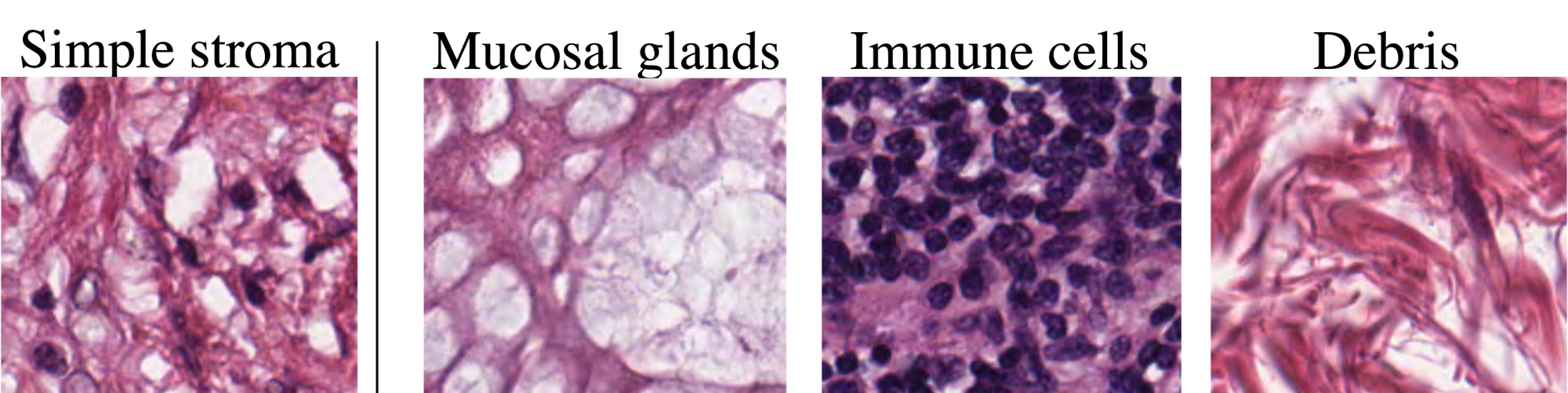}
		\caption{CCH5000 \cite{kather2016multi}}
		\label{fig:data_colo}
	\end{subfigure}%	
	
	\begin{subfigure}{\linewidth}
		\centering
		\includegraphics[width=\linewidth]{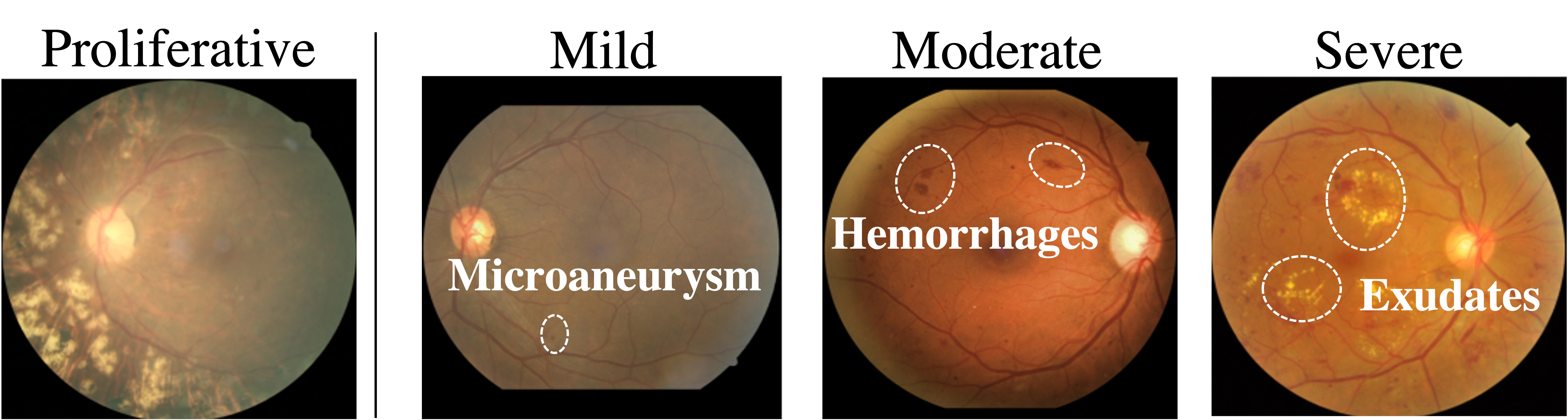}
		\caption{EyePACS \cite{kaggle-diabetic-retinopathy}}
		\label{fig:data_diab}
	\end{subfigure}
	
	\caption{Sample images  illustrating new class (first image) have shared features from old classes.}
	\label{fig:data}
\end{figure}

Existing work on class-incremental continual learning focuses on catastrophic forgetting \cite{mccloskey1989catastrophic,goodfellow2013empirical}, where performance of DNNs degrades significantly on previously learned classes as more classes are being introduced. For the medical domain,
we should in fact go beyond just alleviating forgetting. As shown in our previous example, one observation is that diseases tend to have overlapping features.  Particularly, classification of a new class could depend on features from various old classes. Figure \ref{fig:data}(a) shows the structure of simple stroma that could be described using combination of texture from other tissues. Similarly, the progression of a disease includes earlier clinical symptoms. Figure \ref{fig:data}(b) shows that the proliferative stage of diabetic retinopathy includes clinical signs from mild, moderate and severe stages.  
This motivates us to explore how we can leverage on the previously learned knowledge to acquire new features that allow us to discriminate the expanding set of classes, and possibly even lead to performance improvement on the old disease classes.

There has been limited research on class-incremental continual learning in the medical domain. \citeauthor{li2020continual} \shortcite{li2020continual} introduced an ensemble strategy to update representation of old classes. However, it still lacks in terms of utilizing previously learned features. A Bayesian generative model for medical use cases is proposed by \citeauthor{yang2021continual} \shortcite{yang2021continual} where old classes are represented using statistical distributions. It allows the model to preserve old knowledge, but faces difficulty learning new features due to the fixed pre-trained feature extractor.

In this work, we design a framework that leverages on what has been learned to derive new features for recognizing the cumulative set of classes.
We employ a dynamic architecture that allows  features from the old classes to be reused while expanding the set of feature extractors to learn novel features from the new classes. The proposed architecture has a single low-level feature extractor shared across all classes, which helps to promote utilization of previously learned knowledge. 
We propose a new training strategy that alternates between two objective functions, one focusing on new classes and the other on the old. With this, the classifier can perform well on new classes while still able to maintain, or even improve, its performance on old classes.

We validate our proposed framework  on three publicly available medical datasets.
Empirical results show that our approach outperforms state-of-the-art methods, especially when the dataset is highly skewed and the incremental classes per step is small. Aside from being able to alleviate forgetting, the results show that we are able to utilize newly learned features to better discriminate old classes. 

\section{Related Work}

Research on class-incremental continual learning has largely been concentrated in the natural image domain. These works can broadly be categorized into three approaches: regularization-based, replay-based, and architecture-based. 

Regularization-based approach preserves old knowledge by  using  additional regularization term to penalize changes in previously learned features. Previous works \cite{kirkpatrick2017overcoming,zenke2017continual,aljundi2018memory} estimate the importance of each weight in a previously learned model and apply penalty if there are updates to the important weights. 
Other works use distillation loss to ensure that the features learned of the old classes are preserved \cite{li2017learning,hou2018lifelong,douillard2020podnet,kim2021split}. Another interesting direction introduced in recent work \cite{tao2020topology} is the use of topology-preserving loss to maintain the feature space topology. However, one difficulty faced by approaches in this category is balancing the regularization term such that learning of new classes would not be hindered.

Data replay-based methods  interleaves data related to previous tasks with new data so that the model is `reminded' of the old knowledge.  One way to obtain the old data is by training deep generative models to synthesize `fake' samples that mimics data from previous classes \cite{shin2017continual,wu2018memory,rao2019continual,wang2022continual} but there is the issue of generating realistic data samples. Another way is to directly store a small number of previous class samples to train together with the new data \cite{rebuffi2017icarl,chaudhry2019continual}. 
However, due to the limited memory buffer, there is an imbalance between the small number of old class samples and the relatively large amount of data for new classes.
Attempts to address this issue  include employing a two-stage learning where the classifier is fine-tuned using a balanced dataset \cite{castro2018end} or by correcting the biased weights after training \cite{wu2019large,zhao2020maintaining}.
On the other hand, a few works focus on how best to select data such that old class performance could be maintained. Particularly, \citeauthor{aljundi2019online} \shortcite{aljundi2019online} proposed to store data instances that suffer most by the parameters update of the new model, while \citeauthor{shim2021online} \shortcite{shim2021online} selects samples that best preserve decision boundaries between classes.
Other works propose to parameterize and construct the representational data of each seen class through bi-level optimization \cite{liu2020mnemonics,chaudhry2021using}.

Architecture-based approaches focus on dynamically expanding the network structure and allocate new model parameters to accommodate new information while keeping previously learned parameters fixed to preserve old knowledge \cite{hung2019compacting,mallya2018piggyback,fernando2017pathnet,yoon2017lifelong}.
Most of these methods  use different part of the network  for each task which requires  task identity during inference, but this identity information is usually unavailable. The work by \citeauthor{yan2021dynamically} \shortcite{yan2021dynamically} introduces the notion of  expandable representation learning by adding new branches to the feature extractor of existing network to learn novel concepts for the  incoming data while fixing the weights of old branches to preserve previously learned features. However, there is minimal exploitation of old knowledge since each feature extractor branch is independent of each other.

\section{Proposed Approach}

\begin{figure*}[!t]
	\centering
	\includegraphics[width=0.87\linewidth]{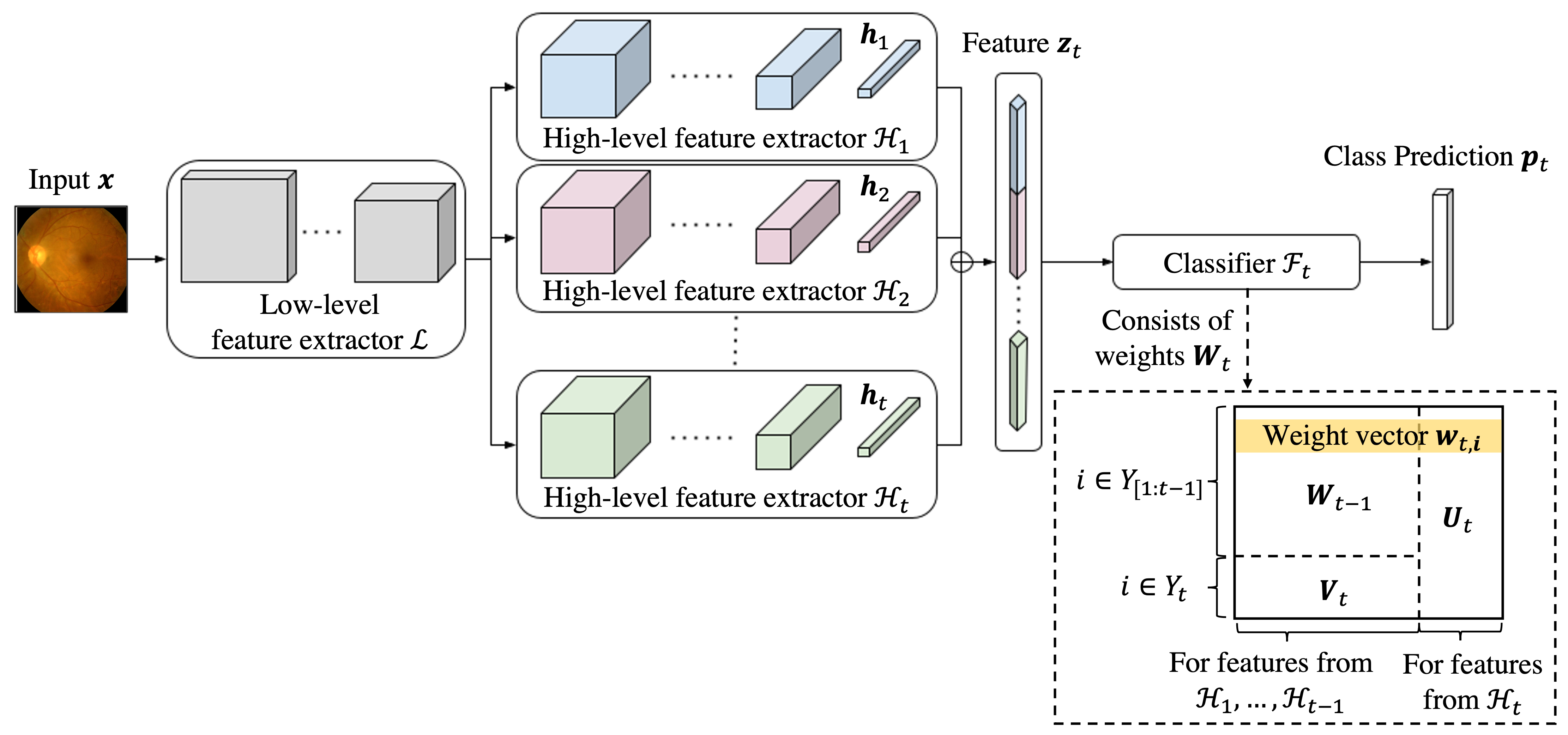}
	\caption{Model architecture at incremental step $t$, consisting of low-level feature extractor $\mathcal{L}$, high-level feature extractors $\{\mathcal{H}_1,\dots,\mathcal{H}_t\}$ and unified classifier $\mathcal{F}_t$. Low-level features learned are shared across all tasks and a new high-level feature extractor $\mathcal{H}_t$ is added each step $t$. The classifier $\mathcal{F}_t$ consists of weights with vectors $\bm{w}_{t,i}$ for all classes $i \in Y_{[1:t]}$.}
	\label{fig:arch}
\end{figure*}

In class-incremental continual learning, the model is required to learn from a stream of data. At each incremental step $t \in [1..T]$, let $Y_t$ be the set of new classes and $D_t$ be the dataset comprising samples $(\bm{x}, y)$, where $\bm{x}$ denotes an input image and $y \in Y_t$ is the corresponding label. The goal is to maximize the overall classification accuracy of all seen classes up to step $t$, i.e. $Y_{[1:t]} = \cup_{i=1}^t Y_i$. 

Our proposed framework  leverages previously learned features to learn new classes. We utilize a dynamically expanding network to accommodate new features without compromising old ones. To maintain performance on previously seen classes, regularization loss and data replay strategy are also employed. Further, to handle the highly skewed class distribution, we use cost-sensitive learning \cite{ting2000comparative} and assign higher penalty to samples from under-represented old classes. 

\subsection{Model Architecture}
\label{sec:arch}

Figure \ref{fig:arch} shows the details of our proposed dynamic architecture at incremental step $t$.
There are three main components: a low-level feature extractor, a set of high-level feature extractors, and a unified classifier. The low-level feature extractor $\mathcal{L}$, parameterized by $\bm{\theta}_{\mathcal{L}}$, is shared throughout the learning process. 
Each high-level feature extractor $\mathcal{H}_k$ has the same architecture that receives processed input from $\mathcal{L}$ and builds upon the low-level features to learn discriminating features for the new set of classes. A new high-level feature extractor $\mathcal{H}_k$, parameterized by $\bm{\theta}_{\mathcal{H}_k}$, is added to the model at each step $t$. 
By allowing the low-level feature extractor to be shared across various tasks, our approach encourages the reuse of  similar features by the high-level feature extractors. The output is a vector $\bm{h}_k$ of dimension $d$. To alleviate forgetting, we freeze the set of old parameters $\{\bm{\theta}_{\mathcal{L}},\bm{\theta}_{\mathcal{H}_1},\dots,\bm{\theta}_{\mathcal{H}_{t-1}}\}$.

The unified classifier $\mathcal{F}_{t}$ is  a single layer with weight matrix $\bm{W}_t$ comprising of vectors $\bm{w}_{t,i}$ for each class $i \in Y_{[1:t]}$. These vectors are derived from the weight matrix of previous step $t-1$ by expanding row-wise to accommodate new classes $Y_t$ and column-wise to include new features from $\mathcal{H}_t$.  Let $\bm{W}_{t-1}$ be the weight matrix comprising of vectors $\bm{w}_{t-1,i}$ of all seen classes  $i \in Y_{[1:t-1]}$. Then the matrix $\bm{W}_{t}$ is given by
$\bm{W}_t=\big[[\bm{W}_{t-1}\circ\bm{V}_{t}] ; \bm{U}_t\big]$, where $\bm{U}_{t}$ is a matrix with weight vectors corresponding to features from $\mathcal{H}_t$ for all classes in $Y_{[1:t]}$ and $\bm{V}_t$ is a matrix of feature  weights  from $\mathcal{H}_1$ up to $\mathcal{H}_{t-1}$ for new classes in $Y_t$. 

Suppose $\bm{z}_t = [\bm{h}_1 \circ \bm{h}_2 \circ \cdots \circ \bm{h}_t]$ is the concatenated outputs from all high-level feature extractors for an input image $\bm{x}$. Then the probability of input $\bm{x}$ belonging to class $i$ can be estimated as follows: 
\begin{equation}
	p_i(\bm{z}_t)= \frac{e^{\eta\cdot\text{sim}(\bm{z}_t,\bm{w}_{t,i})}}{\sum_{j\in Y_{[1:t]}}e^{\eta\cdot\text{sim} (\bm{z}_t, \bm{w}_{t,j})}}
\end{equation} where $\eta$ is a learnable scalar and sim$(\cdot)$ is the cosine similarity between two vectors \cite{hou2019learning}.

\subsection{Training Procedure}
\label{sec:train}

Following the data replay strategy, our training samples $S_t$ at each incremental step $t>1$ consists of the incoming dataset $D_t$ and a limited number of samples from each seen classes  in memory  $M_{t-1}$, that is, $S_t = M_{t-1} \cup D_t$. Our goal is to optimize the new model parameters  $\bm{\theta}_{\mathcal{H}_t}, \bm{V}_t, \bm{U}_t$ using $S_t$ while keeping old parameters $\bm{\theta}_{\mathcal{L}},\bm{\theta}_{\mathcal{H}_1}, \dots, \bm{\theta}_{\mathcal{H}_{t-1}}, \bm{W}_{t-1}$ fixed to preserve previous knowledge. 

We design two objectives $L^{new}$ and $L^{old}$ and use them \emph{alternately} during each training step. The objective $L^{new}$ focuses on learning discriminative features of the new incoming set of data and has four components namely, classification loss, auxiliary loss, distillation loss and margin loss:
\begin{equation}
	\label{eq:l_new}
	L^{new}=L_{class} + \lambda_1 L_{aux} + \lambda_2 L_{dist} + \lambda_3 L_{marg}
\end{equation}
where $\lambda_1$,  $\lambda_2$ and $\lambda_3$ are hyper-parameters for balancing the losses.

On the other hand, $L^{old}$  focuses on the under-represented old classes by assigning higher penalty for the misclassified samples and does not  require auxiliary loss. We have 
\begin{equation}
	\label{eq:l_old}
	L^{old}=L_{class} + \lambda_4 L_{dist} + \lambda_5 L_{marg}
\end{equation}
where $\lambda_4$ and $\lambda_5$ are hyper-parameters.

To prevent the model from neglecting or compromising the newly learned features when optimising $L^{old}$, we freeze the weights $\bm{\theta}_{\mathcal{H}_t}$ and $\bm{U}_{t}$. This preserves the new knowledge learned and the classifier can make use of them to refine the decision boundary between the old and new classes. Details of each loss component are given below.

\paragraph{Classification Loss} 
Instead of the commonly used cross-entropy loss where serious forgetting would occur due to the imbalance between the new and old classes \cite{rebuffi2017icarl}, we  adopt the class-balanced focal loss \cite{cui2019class}. Our classification is given by: 
\begin{equation}
	L_{class}=\mathop{\mathbb{E}}_{(\bm{x},y)\sim S_t}\left[-\frac{1-\beta}{1-\beta^n} \Big(1-p_y(\bm{z}_t)\Big)^\gamma \log\Big(p_y(\bm{z}_t)\Big)\right]
\end{equation} 
where $n$ is the number of samples in $S_t$ with class label $y$. 

By assigning higher penalty on old classes, this loss takes into account class imbalance when learning features to better differentiate old classes from the new. This avoids discarding samples valuable for learning an accurate decision boundary. Note that the  hyper-parameters $\beta$ and $\gamma$ used for this loss in $L^{old}$ are different from those in $L^{new}$ so that misclassification of old class samples are penalized more under the former.

\paragraph{Auxiliary Loss}  
To learn the discriminating features of the new classes, we introduce an auxiliary loss. It also uses the class-balanced focal loss but focuses only on features extracted by $\mathcal{H}_t$, that is, $\bm{h}_t$, as well as the corresponding weight vector $\bm{u}_{t,y}$ for the class $y$ as follows: \begin{equation}
	L_{aux}=\mathop{\mathbb{E}}_{(\bm{x},y)\sim S_t}\left[-\frac{1-\beta}{1-\beta^n} \Big(1-p_y(\bm{h}_t)\Big)^\gamma \log\Big(p_y(\bm{h}_t)\Big)\right]
\end{equation}
where 
\begin{equation}
	p_y(\bm{h}_t)= \frac{e^{\text{sim}(\bm{h}_t,\bm{u}_{t,y})}}{\sum_{i \in Y_{[1:t]}}e^{\text{sim}(\bm{h}_t\,\bm{u}_{t,i})}}
\end{equation}

Note that we use existing parameters in the classifier $\mathcal{F}_t$ to learn a better decision boundary in the new feature dimension, unlike previous work \cite{yan2021dynamically} which introduces an additional auxiliary classifier. 

\paragraph{Distillation Loss}  
This regularization term is designed to alleviate forgetting by transferring knowledge from the old model to the new. Since our architecture is designed to freeze previously learned features, we use logits-level distillation loss \cite{rebuffi2017icarl} by minimizing Kullback–Leibler divergence \cite{kullback1951information} between the probabilities of old classes $Y_{[1:t-1]}$ predicted by the model at previous step as follows:  
\begin{equation}
	L_{dist}=\!\mathop{\mathbb{E}}_{(\bm{x},y)\sim S_t}\!\! \left [ \Big|Y_{[1:t-1]}\Big|\!\!\sum_{i \in Y_{[1:t-1]}}\!\!\! p_i(\bm{z}_{t-1})\log\frac{p_i(\bm{z}_{t-1})}{p_i(\bm{z}_{t})} \right ]
\end{equation}

We weight the loss to take into account that the need to preserve previously learned knowledge varies with the number of old classes.

\paragraph{Margin Loss}  
Since the training set is dominated by new classes, the predictions may be biased towards them. To reduce the bias, we use margin ranking loss \cite{hou2019learning} such that samples extracted from memory $M_{t-1}$ have well-separated ground truth old class from all the new classes with margin of at least $m$. Given a training sample $(\bm{x},y)$, let $K$ be the set of corresponding new classes with the $k$ highest predicted confidence. Then the loss is defined as:
\begin{multline}
	L_{marg}=\mathop{\mathbb{E}}_{(\bm{x},y)\sim M_{t-1}}\Bigg [\sum_{i \in K} \max\Big(  \text{sim}(\bm{z}_t,\bm{w}_{t,i}) - \\ \text{sim}(\bm{z}_t,\bm{w}_{t,y}) + m, 0 \Big) \Bigg ]
\end{multline} 

\begin{algorithm}[t]
	\caption{Proposed Training Procedure}\label{alg}
	\begin{algorithmic}[1]
		\Require Initialized model parameters $\mathcal{L}$, $\mathcal{H}_1$, $\mathcal{F}_1$
		\Statex ~~\quad\quad Number of incremental steps $T$
		\Statex ~~\quad\quad Datasets at each step  \{$D_1$,$\cdots$,$D_{T}$\}
		\Statex ~~\quad\quad Number of training epochs $N$
		
		\State Train $\mathcal{L}, \mathcal{H}_1, \mathcal{F}_{1}$ with $L_{class}$ using $D_1$ for $N$ epochs  %\Comment{Training for first task}
		\For{$t \leftarrow 2, \dots, T$} %\Comment{Training for step $t$}
		\State Expand architecture with $\mathcal{H}_t$ and $\mathcal{F}_{t}$ 
		\State Construct memory $M_{t-1}$
		\State $S_t \leftarrow M_{t-1} \cup D_t$ 
		%\State Transfer weights $\bm{W}_{t-1}$ to $\bm{W}_{t}$ 
		\Repeat   %\Comment{Alternate objectives}
		\State Train with $L^{new}$ from Eq. (\ref{eq:l_new}) using $S_t$ 
		\State Train with $L^{old}$ from Eq. (\ref{eq:l_old}) using $S_t$ 
		\Until epoch = $N$
		\EndFor
	\end{algorithmic}
\end{algorithm}
Algorithm \ref{alg} gives the details of our proposed training approach. In Line 1, $D_1$ is used to train the low-level feature extractor $\mathcal{L}$, high-level feature extractor $\mathcal{H}_1$, and classifier $\mathcal{F}_1$ with only classification loss as there is no old knowledge to preserve. For each subsequent step $t$, the architecture is expanded with a new high-level feature extractor $\mathcal{H}_t$ and a larger classifier $\mathcal{F}_t$ (Line 3). Line 4 selects old samples for data replay $M_{t-1}$ while Line 5 merges the samples with $D_t$ to obtain training dataset $S_t$. The training procedure (Lines 6-9) is repeated by first optimizing $L^{new}$, followed by $L^{old}$.

\section{Experiments}
\label{sec:exp}
\paragraph{Datasets and Settings}
\label{sec:setting}
We follow the protocol where we train  the model using half of the classes and split the remaining classes evenly for training in each incremental step \cite{hou2019learning}. We use the following datasets and settings in our experiments:
\begin{itemize}
	\item CCH5000 \cite{kather2016multi}: This dataset consists of histological images with each belonging to one of 8 tissue categories that represents textures in of human colorectal cancer. It has a uniform class distribution of 625 images per class, which we randomly select 20\% from each for testing. We use 4 classes to train an initial model, and the remaining are split into groups of  1 and 2 classes for the two respective incremental learning settings.
	
	\item EyePACS \cite{kaggle-diabetic-retinopathy,cuadros2009eyepacs}: This Kaggle dataset consists of 35125  retinal images that have been graded  for the severity of diabetic retinopathy (DR).	There are 5 classes: no DR, mild DR, moderate DR, severe DR and proliferative DR. It has highly skewed distribution, with  73\%  of the images having no DR and only 2\% having severe DR. We use the first 3 classes  to train an initial model and incrementally learn the remaining two classes. 
	
	\item HAM10000 \cite{tschandl2018ham10000,DVN/DBW86T_2018}: This dataset is a collection of 7   types of pigmented skin lesions. There are 10015 images, from which 20\% is randomly selected for evaluation. The class distribution is highly uneven, with number of images per class varying from 115 to 6705. We train an initial model on 3 classes, and use the remaining 4 classes for incremental learning under two settings where 1 or 2 classes are introduced at each step respectively. 	
\end{itemize} 

For each setting, the experiments are repeated on three random class order. All the training images are augmented with random flipping and cropping. 
Following previous work \cite{rebuffi2017icarl}, we use the herd selection strategy \cite{welling2009herding} to select a fixed number of 20 samples per class for data replay in all experiments. 

\paragraph{Baselines}
We compare our approach with the following baselines: (a) iCaRL \cite{rebuffi2017icarl} alleviates forgetting  using logits-level distillation loss; (b)
UCIR \cite{hou2019learning} preserves previously learned knowledge by fixing the weight vectors of old classes and uses feature-level distillation loss  as well as margin loss in their solution; (c)
PODNet \cite{douillard2020podnet} prevents forgetting with spatial-based distillation loss and uses  two-stage learning to address class imbalance; (d) DER \cite{yan2021dynamically} uses dynamically expandable representation to handle new classes without forgetting and also two-stage learning for the class imbalance issue.

\paragraph{Evaluation Metrics}
After each incremental step $t$, the model is evaluated only on all the classes seen so far. We denote the accuracy of classifying samples with classes in $Y_i$ using model trained at step $t$ as $A^i_{t}$. 
The overall accuracy is computed as $Acc = \frac{1}{T}\sum_{t=1}^{T} \left[ \frac{1}{t} \sum_{i=1}^{t} A^i_{t}\right]$. In addition to the $Acc$ metric, we also quantify the amount of forgetting of old classes as the difference between accuracy at current step and the maximum obtained before that. This is given by $Fgt = \frac{1}{T}\sum_{t=1}^{T}\left[\frac{1}{t-1}\sum_{i=1}^{t-1}\max_{j=1}^{t-1}(A^i_{j} -A^i_{t})\right]$.
To determine the significance in performance improvement of our approach over the next-best-performing baseline, we run paired $t$-test on the metrics.

\paragraph{Implementation Details}
\label{implementation}
All methods are implemented in PyTorch \cite{paszke2019pytorch} and trained on NVIDIA V100 GPU with 32GB memory. We adopt ResNet-18 \cite{he2016deep} as the the network backbone and cosine normalization \cite{hou2019learning} in the classifier layer. For our approach, we use the first two residual blocks as the low-level feature extractor and introduce duplicates of the remaining two residual blocks at each incremental step. Backbone of all models are initialized with weights pre-trained on ImageNet \cite{deng2009imagenet} and optimized using SGD optimizer with momentum value of 0.9 and weight decay of 0.0005. 
We use batch size of 32 for CCH5000 and HAM10000, and 128 for EyePACS. 
Details of the hyper-parameters, data pre-processing, data sampling can be found in our implementation code at \url{https://github.com/EvelynChee/LO2LN.git}.

\subsection{Comparative Study}
\label{sec:com}

\paragraph{Results on CCH5000}

Table \ref{tab:res_CCH5000} summarizes the results for the CCH5000 dataset. We see that our proposed approach outperforms all the baselines for both settings in terms of $Acc$ and $Fgt$. Particularly,  it is more advantageous under the setting of 1 new class per step, in which there is an increase of 2.5\% in accuracy and a drop of 1.6\% in forgetting when compared with the next-best-performing baseline. When trained incrementally with 2 new class per step, our model shows an improvement in accuracy and forgetting of 0.7\% and 0.4\% respectively.

\begin{table}[!b]
	\small
	\centering	
	\begin{tabular}{>{\raggedright\arraybackslash}p{1.1cm}|>{\centering\arraybackslash}p{1.34cm}>{\centering\arraybackslash}p{1.25cm}|>{\centering\arraybackslash}p{1.34cm}>{\centering\arraybackslash}p{1.25cm}}
		\toprule[0.1em]
		Setting & \multicolumn{2}{c|}{1 new class per step} & \multicolumn{2}{c}{2 new classes per step}  \\ \hline
		Metric & $Acc$ & $Fgt$ & $Acc$ & $Fgt$  \\ 
		\midrule[0.1em]
%		Joint & 96.9\small$\pm$0.2 & - & 97.2$\pm$0.2 & - \\  \cmidrule{1-5}
		iCaRL & 91.1$\pm$1.8 &  9.0$\pm$3.3 & 93.0$\pm$0.2 & 6.8$\pm$1.0   \\
		UCIR  & 92.0$\pm$1.0 &  5.5$\pm$2.6 & 93.9$\pm$0.3 & 4.4$\pm$0.9 \\
		PODNet  & 89.2$\pm$0.5 & 6.0$\pm$1.2 & 92.0$\pm$0.3 & 5.2$\pm$0.4  \\
		DER & 91.0$\pm$1.7 &  5.6$\pm$1.9 & 93.0$\pm$0.5 & 6.4$\pm$1.4 \\
		Ours & \bf{94.5$\pm$0.8}$^*$ & \bf{3.9$\pm$2.0} & \bf{94.6$\pm$0.4}$^*$ & \bf{4.0$\pm$0.8}  \\  		
		\bottomrule[0.1em]             
	\end{tabular} 	
	\raggedright
	$^*$Statistically significant improvement with $p$-value$<$0.05   
	\caption{\normalsize Results on CCH5000 over three runs. The upper bound $Acc$ for the setting of 1 and 2 new classes per step, achieved by keeping all previous training data  accessible, are 96.9\% and 97.2\% respectively.}			
	\label{tab:res_CCH5000}	
\end{table}

\begin{figure}[!b]
	\centering
	\begin{subfigure}{\linewidth}
		\centering
		\includegraphics[width=0.98\linewidth]{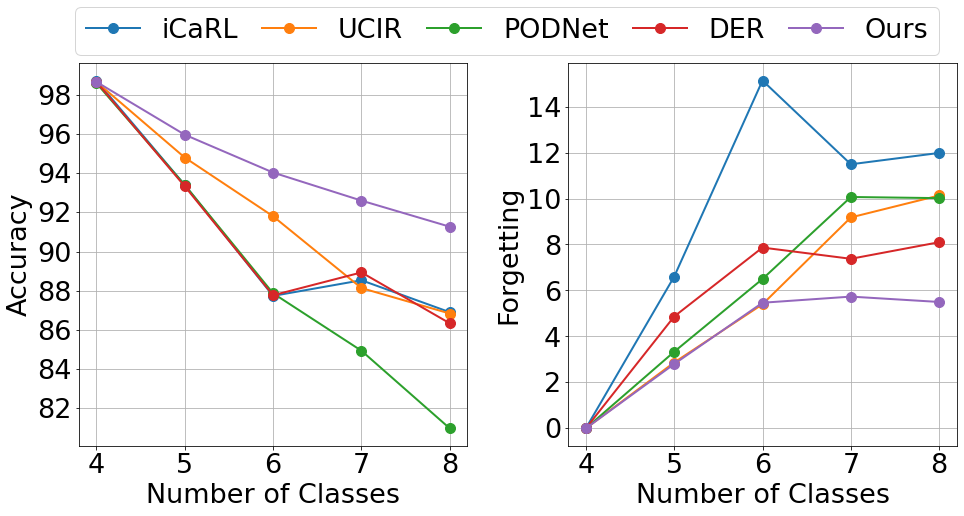}
		\caption{1 new class per step}
	\end{subfigure}%
	
	\begin{subfigure}{\linewidth}
		\centering
		\includegraphics[width=0.98\linewidth]{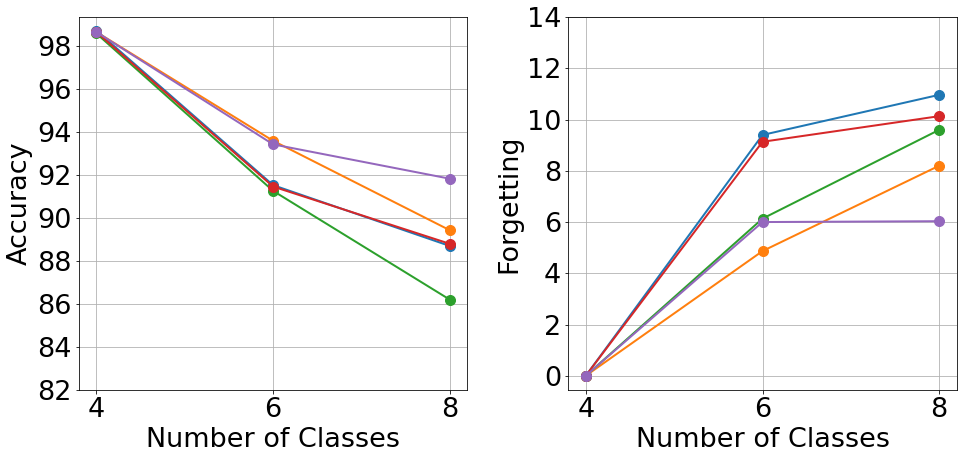}
		\caption{2 new classes per step}
	\end{subfigure}
	\caption{Accuracy and forgetting at each incremental step on CCH5000, averaged across three runs. }
	\label{fig:cch} 
\end{figure}

\begin{figure*}[!t]
	\centering
	\includegraphics[width=0.949\linewidth]{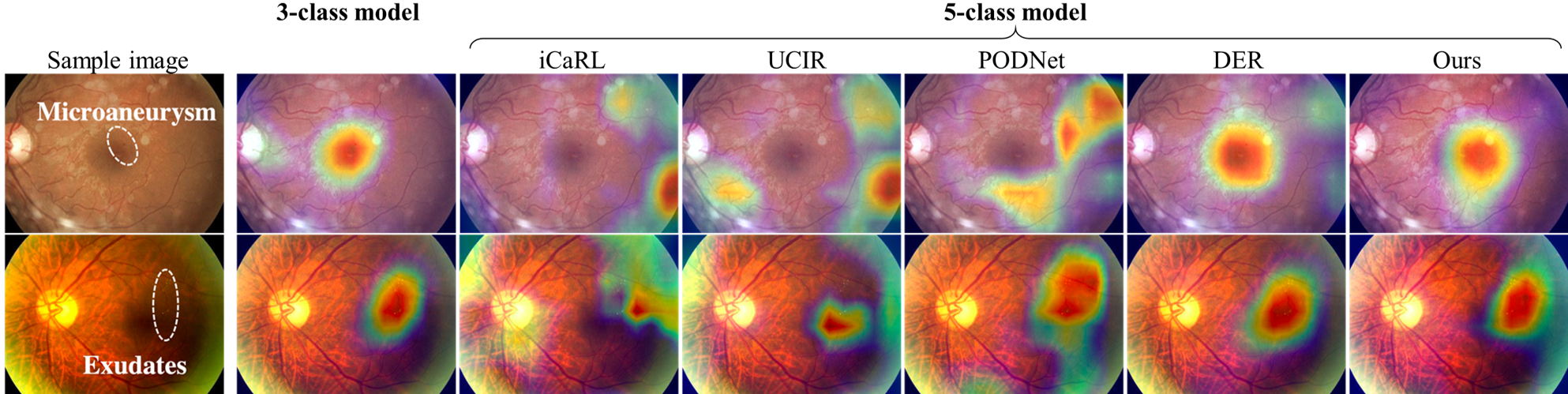}
	\caption{Activation maps on sample EyePACS images with classes seen at $t$=1. The second column is based on the model trained on the first 3 classes, while the remaining have been trained on all 5 classes.}
	\label{fig:kaggle_feat2}
\end{figure*}

Figure \ref{fig:cch} compares the performance of all methods at each incremental step. We see that the gap between our approach and state-of-the-art methods widens as more classes are introduced. This confirms that our approach is effective in learning the new while not forgetting the old. Similarly, a larger improvement is seen for the smaller incremental class setting (1 new class per step) where our method outperforms the nearest baseline in terms of final accuracy by 4.4\% (from 86.9\% to 91.3\%) and final forgetting by 2.6\% (from 8.1\% to 5.5\%). As for the setting of 2 new classes per step, the performance gap at the final step are 3.1\% (from 88.7\% to 91.8\%) and 2.2\% (from 8.2\% to 6.0\%) respectively.

\paragraph{Results on EyePACS}

Table~\ref{tab:res_EYEPACS} shows the performance of the various methods on the  EyePACS dataset where the number of new classes per step is 1. 
Our approach achieves significant performance improvement over UCIR where $Acc$ is boosted by 11.7\% and $Fgt$ is reduced  by 15.6\%.  

\begin{table}[!t]
	\small
	\centering	
	\begin{tabular}{>{\raggedright\arraybackslash}p{0.005cm}>{\raggedright\arraybackslash}p{1.5cm}|>{\centering\arraybackslash}p{1.8cm}>{\centering\arraybackslash}p{1.8cm}>{\raggedright\arraybackslash}p{0.005cm}}
		\toprule[0.1em]
		& Setting & \multicolumn{2}{c}{1 new class per step} & \\ \hline
		& Metric & $Acc$ &  $Fgt$ & \\ \midrule[0.1em]
		%& Joint & 84.2\small$\pm$2.2 & - & \\  \cmidrule{1-4}
		& iCaRL & 64.4\small$\pm$3.3  &17.8\small$\pm$4.6 & \\
		& UCIR   & 70.2\small$\pm$7.6 & 15.4\small$\pm$11.4 &  \\
		& PODNet  & 63.3\small$\pm$5.4 &22.8\small$\pm$4.9 &  \\
		& DER  & 58.7\small$\pm$9.4 & 30.2\small$\pm$6.9 & \\
		& Ours & \bf{81.9\small$\pm$ 2.5}$^*$ & \bf{-0.2\small$\pm$0.8} & \\  		
		\bottomrule[0.1em]                			
	\end{tabular}	\\
	$^*$Statistically significant improvement with $p$-value$<$0.1 
	\caption{\normalsize Results on EyePACS over three runs. The upper bound $Acc$ for the setting of 1 new class per step, achieved by keeping all previous training data accessible, is 84.2\%.}		
	\label{tab:res_EYEPACS}			
\end{table}

When we compare Tables 1 and 2, we see a general drop in accuracy and increased forgetting  for all the baseline methods. This is because unlike CCH5000, EyePACS has a highly imbalanced class distribution, making continually learning in such skewed datasets more challenging as predictions could easily bias towards new classes that are heavily over-represented. In spite of this, our approach is able to overcome such issue with the use of alternate training objectives that focus on the under-represented old classes. 

\begin{figure*}[!t]
	\centering
	\begin{subfigure}{0.5\linewidth}
		\centering
		\includegraphics[width=0.83\linewidth]{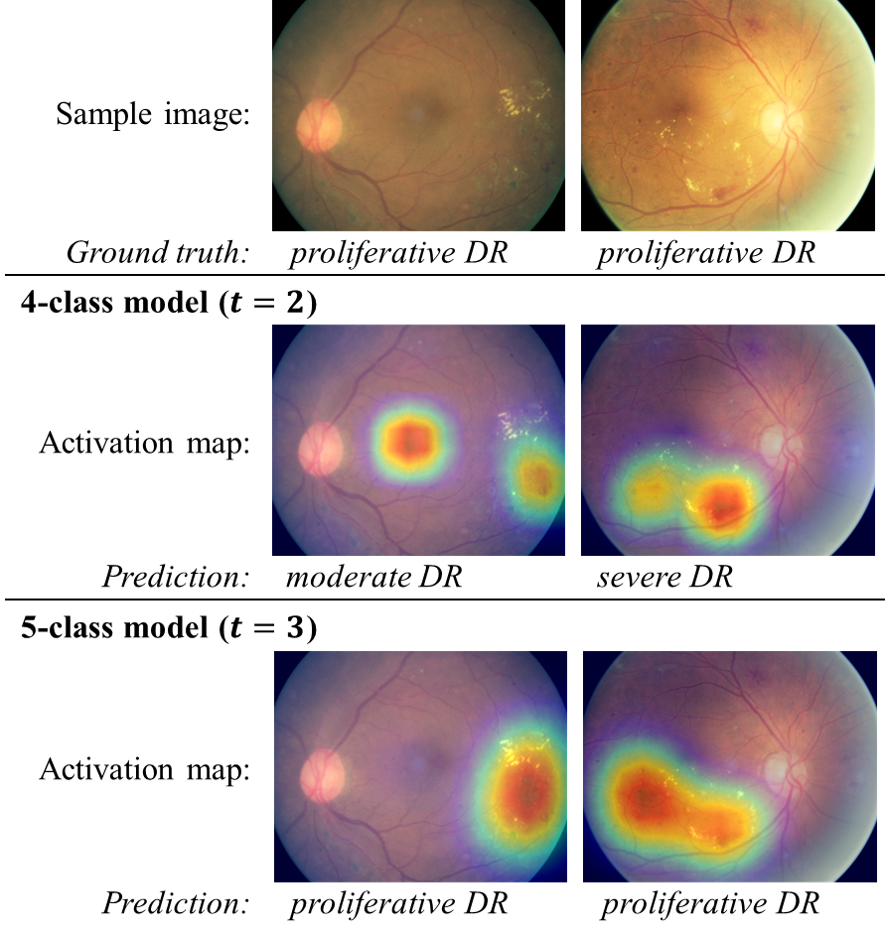}
		\caption{EyePACS}
		\label{fig:kaggle_feat} 
	\end{subfigure}%	
	\begin{subfigure}{0.5\linewidth}
		\centering
		\includegraphics[width=0.83\linewidth]{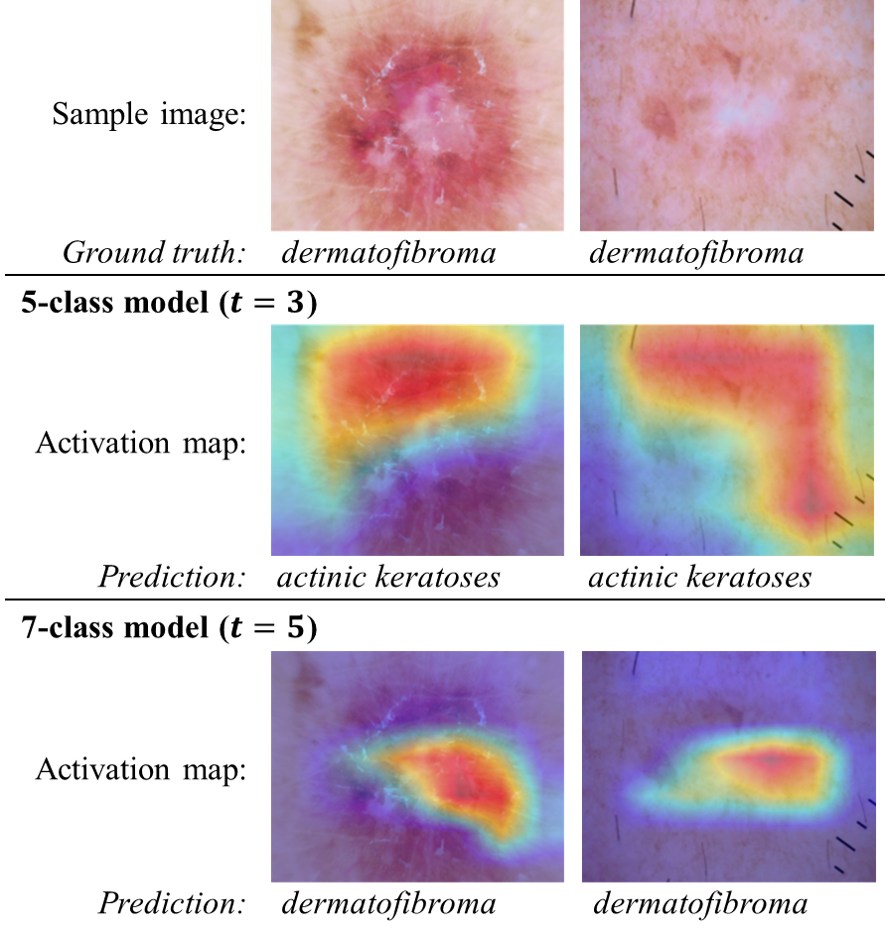}
		\caption{HAM10000}
		\label{fig:ham_feat} 
	\end{subfigure}
	\caption{Activation maps and predictions on sample images using our model trained at different incremental steps $t$. }	
	\label{fig:feat_learn} 
\end{figure*}

Figure \ref{fig:kaggle_feat2} shows the class activation maps obtained using Grad-CAM \cite{selvaraju2016grad} for the model trained on the first three classes (3-class model) as well as those for the various methods after continual learning all the five classes (5-class model).
We observe that the activation maps of DER and our proposed approach have the most overlapped regions with that of the 3-class model, indicating the effectiveness of dynamically expandable representation in preserving the old features.
However,  DER wrongly predicted both samples of moderate DR as proliferative DR. This is due to the two-stage learning approach used in DER, where the classifier is re-initialized and fine-tuned using a balanced dataset in the second stage. 
Discarding the previously learned class weight vectors corresponding to the old features leads to loss in accuracy and increased forgetting.

We also note that our approach is the only one to attain a negative forgetting, suggesting that we are able to utilize newly learned knowledge to improve the classification of previous classes.   Figure \ref{fig:kaggle_feat} shows two samples that are misclassified by the 4-class model (already seen classes no DR, moderate DR, severe DR, and proliferative DR) at $t$=2 and their corresponding activation maps.
After the class mild DR is introduced at $t$=3, we observe that the 5-class model focuses better on the relevant clinical symptoms (small bright exudates spots) and gives the correct predictions. This suggests that our model realizes the importance of this symptom due to its absence in milder cases and thus, updates its knowledge on old classes to achieve negative forgetting.

\paragraph{Results on HAM10000}

\begin{table}[!t]
	\small
	\centering	
	\begin{tabular}{>{\raggedright\arraybackslash}p{1.08cm}|>{\centering\arraybackslash}p{1.34cm}>{\centering\arraybackslash}p{1.185cm}|>{\centering\arraybackslash}p{1.34cm}>{\centering\arraybackslash}p{1.345cm}}
		\toprule[0.1em]
		Setting & \multicolumn{2}{c|}{1 new class per step} & \multicolumn{2}{c}{2 new classes per step}  \\ \hline
		Metric & $Acc$ & $Fgt$ & $Acc$ & $Fgt$  \\ \midrule[0.1em]
		%Joint & 89.3$\pm$2.1 & - &  89.1$\pm$2.2 & - \\ \cmidrule{1-5}
		iCaRL  & 68.3\small$\pm$2.8 & 25.3\small$\pm$4.5 &  76.3\small$\pm$3.1 & 20.1\small$\pm$12.6     \\
		UCIR  &  74.1\small$\pm$3.1  & 16.3\small$\pm$9.1 & 79.1\small$\pm$1.4 & 16.8\small$\pm$9.5 \\
		PODNet  & 66.3\small$\pm$2.3 & 17.3\small$\pm$4.8 & 75.6\small$\pm$2.2 & 20.5\small$\pm$2.1 \\
		DER &  66.9\small$\pm$4.5  &24.7\small$\pm$4.8 & 76.2\small$\pm$2.8 & 24.8\small$\pm$10.9 \\
		Ours & \bf{78.1\small$\pm$3.4}$^*$ &\bf{10.1\small$\pm$3.9}  &  \bf{82.0\small$\pm$1.3}$^*$ & \bf{12.8\small$\pm$3.3}\\ 
		\bottomrule[0.1em]                			
	\end{tabular}	
	\raggedright
	$^*$Statistically significant improvement with $p$-value$<$0.05   
	\caption{\normalsize Results on HAM10000 over three runs. The upper bound $Acc$ for the setting of 1 and 2 new classes per step, achieved by keeping all previous training data  accessible, are 89.3\% and 89.1\% respectively.}			
	\label{tab:res_HAM10000}
\end{table}

We also compare the performance of the various methods on HAM10000. Table \ref{tab:res_HAM10000} shows that our approach outperforms the baselines by a wide margin for both $Acc$ and $Fgt$. Again, the improvement over the next best method is more significant under the setting of 1 new class per step, where there is an increase of 4.0\% in accuracy and a drop of 6.2\% in forgetting.

Figure \ref{fig:ham_feat} shows the activation maps of two dermatofibroma samples characterized by central white patches \cite{zaballos2008dermoscopy}. At $t$=3, the 5-class model receives and is trained with new data on this class but its activation maps for these two samples are outside the white patches, indicating that the model is not focusing on the correct features and hence misclassifies. In contrast, the 7-class model at $t$=5 is able to focus on the relevant regions and improves its predictions on the previously learned class.

\subsection{Ablation Study}
\label{sec:abl}

We analyze the effect of different loss components in our training objectives.
Table \ref{tab:abl} shows the results of ablation study using CCH5000 under the incremental setting of 1 new class per step. 
Besides the metric $Acc$, we also report the average accuracy on samples of classes newly learned at each incremental step $t$ (i.e., $Acc_{new}=\frac{1}{T}\sum_{t=1}^{T}A^t_{t}$) and those of classes introduced at $t$=1 (i.e., $Acc_{old}=\frac{1}{T}\sum_{t=1}^{T}A^1_{t}$) since they depict  the model's ability to learn new concepts and preserve old knowledge respectively. 

With all the loss components incorporated, our approach achieves the highest $Acc$ and the best balance between $Acc_{new}$ and $Acc_{old}$. 
Removing the margin loss results in the steepest drop in $Acc$ and $Acc_{old}$, indicating its role in alleviating forgetting. However, it has also resulted in the highest $Acc_{new}$, which demonstrates its impact on learning of new knowledge.
As for $L^{old}$, there is a drop in $Acc_{old}$ when it is not used, suggesting the effectiveness of our alternating objective functions in preserving old knowledge.

\begin{table}[!t]
	\small
	\centering
	\begin{tabular}{l|>{\centering\arraybackslash}p{1.5cm}>{\centering\arraybackslash}p{1.5cm}>{\centering\arraybackslash}p{1.5cm}}
		\toprule[0.1em]
		& $Acc$ & $Acc_{new}$  & $Acc_{old}$    \\
		\midrule[0.1em]
		
		All & \bf{94.5\small$\pm$0.8} & 96.6\small$\pm$0.9 & 94.0\small$\pm$0.7\\
		Without $L^{old}$ & 93.8\small$\pm$1.1 & 97.9\small$\pm$1.0 & 92.6\small$\pm$1.1  \\
		Without $L_{aux}$ & 94.4\small$\pm$0.7 & 96.4\small$\pm$0.6 & 93.7\small$\pm$0.9 \\
		Without $L_{dist}$ & 93.4\small$\pm$0.4 & 96.5\small$\pm$0.6 & 92.4\small$\pm$0.6 \\
		Without $L_{marg}$ & 91.8\small$\pm$1.7 & 98.9\small$\pm$0.5 & 89.3\small$\pm$1.5  \\
		%Without new $\mathcal{H}_t$ & 91.9\small$\pm$1.7 & 96.3\small$\pm$1.7 & 90.6\small$\pm$1.3  \\
		\bottomrule[0.1em]             
	\end{tabular}
	\caption{\normalsize Ablation study on effects of each loss component using CCH5000 over three runs.} 
	\label{tab:abl}
\end{table}

Aside from the loss components, we also analyze the effect of dynamically expanding the network. We train a model using the proposed objective functions but without adding a new high-level feature extractor at each incremental step. The $Acc$, $Acc_{new}$ and $Acc_{old}$ obtained are 91.9\%, 96.3\% and 90.6\%, which  corresponds to a drop of 2.6\%, 0.3\% and 3.4\% respectively. The relatively large drop in $Acc_{old}$ indicate its importance in preserving knowledge.

\section{Conclusion}
\label{sec:lim}

In this paper, we proposed a class-incremental continual learning  framework for the medical domain that leverages on previously learned features to acquire new knowledge. By utilizing a dynamic architecture with expanding representations, it is able to retain old features while learning new ones. We have achieved a good balance in the performance of old and new classes by alternating the training of the model using two objectives, with one focused on learning from new incoming data while the other emphasizing the under-represented old classes. Experiment results on three medical imaging datasets, including those with highly skewed distribution, have demonstrated the effectiveness of our proposed approach over state-of-the-art baselines.
Future work includes investigating the need to expand the model when new classes are introduced and developing a metric to quantify the contribution of adding a feature extractor branch.

\section{Acknowledgments}
This research is supported by the National Research Foundation Singapore under its AI Singapore Programme (Award Number: AISG-GC-2019-001-2A).

% Use \bibliography{yourbibfile} instead or the References section will not appear in your paper
\bibliography{ref}

\end{document}